\documentclass{article}
\usepackage{ijcai24}

\usepackage{times}
\usepackage{soul}
\usepackage{url}
\usepackage[hidelinks]{hyperref}
\usepackage[utf8]{inputenc}
\usepackage[small]{caption}
\usepackage{graphicx}
\usepackage{amsmath}
\usepackage{amssymb}
\usepackage{amsthm}
\usepackage{booktabs}
\usepackage{algorithm}
\usepackage{algorithmic}
\usepackage{multirow}
\usepackage{subfigure}
\usepackage{tabularx}
\usepackage{subcaption}
\usepackage[switch]{lineno}

\title{FairGT: A Fairness-aware Graph Transformer}

\author{
Renqiang Luo$^1$
\and
Huafei Huang$^1$\and
Shuo Yu$^{1,}$\thanks{Corresponding author.}\and
Xiuzhen Zhang$^2$\And
Feng Xia$^2$\\
\affiliations
$^1$Dalian University of Technology, China\\
$^2$RMIT University, Australia\\
\emails
\{lrenqiang, hhuafei\}@outlook.com,
\{shuo.yu, f.xia\}@ieee.org,
xiuzhen.zhang@rmit.edu.au
}

\begin{document}

\maketitle

\begin{abstract}
The design of Graph Transformers (GTs) generally neglects considerations for fairness, resulting in biased outcomes against certain sensitive subgroups.
Since GTs encode graph information without relying on message-passing mechanisms, conventional fairness-aware graph learning methods cannot be directly applicable to address these issues.
To tackle this challenge, we propose FairGT, a Fairness-aware Graph Transformer explicitly crafted to mitigate fairness concerns inherent in GTs.
FairGT incorporates a meticulous structural feature selection strategy and a multi-hop node feature integration method, ensuring independence of sensitive features and bolstering fairness considerations.
These fairness-aware graph information encodings seamlessly integrate into the Transformer framework for downstream tasks.
We also prove that the proposed fair structural topology encoding with adjacency matrix eigenvector selection and multi-hop integration are theoretically effective.
Empirical evaluations conducted across five real-world datasets demonstrate FairGT's superiority in fairness metrics over existing graph transformers, graph neural networks, and state-of-the-art fairness-aware graph learning approaches.
\end{abstract}

\section{Introduction}
% Graph Neural Networks (GNNs), employing message-passing, have demonstrated significant success across various domains such as healthcare \cite{yu2022pandora}, traffic forecasting \cite{hou2023missii}, and social media mining \cite{zehmakan2023why}.
% However, GNNs are limited by their reliance on message-passing mechanism \cite{wu2023kdlgt}.
% In contrast, Graph Transformers (GTs) empower each node within a graph to directly attend to all other nodes, facilitating the aggregation of information from arbitrary nodes \cite{liu2023gapformer}.
% With the incorporation of global attention and long-range interactions for nodes, GTs address the limitations of GNNs, effectively handling issues like over-smoothing \cite{guo2023contranorm}, over-squashing \cite{he2023generalization}, and long-range dependencies \cite{zhang2022hierarchical}.
Graph Transformers (GTs) incorporates global attention and facilitates long-range interactions among nodes \cite{zhu2023hierarchical}, thus effectively addressing challenges of Graph Neural Networks (GNNs) (e.g., over-smoothing \cite{guo2023contranorm} and over-squashing \cite{he2023generalization}) and handling long-range dependencies \cite{zhang2022hierarchical}.
Despite their success, most of the GTs inevitably overlook the bias in the graph data, which leads to discriminatory predictions towards certain sensitive subgroups, such as gender, age, nationality, and race \cite{caton2023fairness}.
That is to say, the issue of fairness in GTs has become a prominent concern when deploying them in real-world scenarios.
% In order to use GTs for many real world applications, such as salary evaluation system, it is critical for GTs to make fair predictions for people with different sensitive values \cite{cong2023fairsample}.
Effectively employing GTs in practical scenarios, like salary evaluation systems, necessitates generating equitable predictions for individuals with diverse sensitive features.

To quantitatively show the fairness issues exist in GTs, we evaluate one of the most typical fairness-aware GNN methods (i.e., FairGNN \cite{dai2023learning}) and three most popular GTs (i.e., GraphTransformer (GraphTrans), Spectral Attention Network (SAN), and Neighborhood Aggregtation Graph Transformer (NAGphormer)) over a real-world dataset (i.e., NBA), with outcomes presented in Table \ref{tab:background}.
Here, we employ Statistical Parity, i.e., Independence, which is a well adopted notion of fairness assessing the equality of prediction results across diverse demographic groups \cite{dwork2012fairness}.
The experimental setup aligns with relevant studies \cite{dwivedi2020generalization,kreuzer2021rethinking,chen2023nagphormer}, and the higher $\Delta_{\textbf{SP}}$ value corresponds to the lower fairness.
It can be seen that the values of $\Delta_{\textbf{SP}}$ of GTs are much higher than that of fairness-aware GNN, which indicates the existence of fairness issue in GTs.
\begin{table}[H]
  \centering
  \setlength\tabcolsep{4.0pt}
  \footnotesize
  \begin{tabular}{lcccc}
    \toprule
    \textbf{Methods} & \textbf{FairGNN} &\textbf{GraphTrans} & \textbf{SAN} & \textbf{NAGphormer} \\
    \midrule
    $\Delta_{\textbf{SP}}$(\%) $\downarrow$ & $1.32$ & $9.01$ & $29.02$ & $16.74$ \\
    \bottomrule
  \end{tabular}
  \caption{The fairness issue of GTs.}
  \label{tab:background}
  \vspace{-1em}
\end{table}

Numerous efforts have been dedicated to improving fairness in graph learning \cite{xia2021graph,ren2023graph}.
Specifically, fairness-aware graph learning methods have emerged, which primarily focus on preventing the misuse of sensitive features through additional regularizations or constraints. \cite{dong2023fairness}.
FairAC \cite{guo2023fair} alleviates this problem by introducing a sensitive discriminator to regulate each group of neighboring nodes chosen in the sampling process, effectively generating a bias-free graph.
Similarly, FairSample \cite{cong2023fairsample} enhances model fairness by developing a trainable neighbor sampling policy through reinforcement learning, further optimizing it with the incorporation of a regularization objective.

In the realm of GTs, however, there has been a noticeable lack of exploration concerning fairness considerations.
The distinctive architecture of GTs, which enables direct node-to-node interactions instead of relying on neighbor information, poses a unique challenge for fairness-aware methods based on the message-passing mechanism \cite{wu2023kdlgt}.
Simply removing sensitive features may not sufficiently enhance fairness, as the correlation between these features and other factors could still introduce bias \cite{mehrabi2021a}.
As a result, current fairness-aware solutions for graph learning methods cannot be readily applied to tackle the unique fairness issues associated with GTs.

In this work, we present a Fairness-aware Graph Transformer (called FairGT\footnote{The source codes and detailed proofs are available at https://github.com/yushuowiki/FairGT.}), which enhances the independence of sensitive features during the training process and thus improving fairness.
Firstly, FairGT employs an efficient strategy by selecting the eigenvector of the adjacency matrix as the fair structural topology encoding.
This departure from the common practice in GTs \cite{dwivedi2020generalization,kreuzer2021rethinking,chen2023nagphormer}, which typically involves utilizing eigenvectors corresponding to the $s$ smallest non-trivial eigenvalues of the Laplacian matrix, contributes to its efficiency.
Secondly, we introduce a fairness-aware node feature encoding method to keep the independence of sensitive features.
Furthermore, we perform a theoretical analysis to evaluate the correlation between the sensitive feature and the two aforementioned encodings introduced by FairGT.
%We then present the theoretical analysis to substantiate the effectiveness of these encodings.

FairGT leverages these two fairness-aware graph information encodings (i.e., structural topology and node feature) as inputs.
In structural topology encoding, our alternative approach utilizes eigenvectors corresponding to the $t$ largest magnitude eigenvalue of the adjacency matrix, providing a fairer representation of the structural topology.
This encoding maintains original sensitive feature distribution.
Furthermore, we construct a sensitive feature complete graph.
Based on this graph, FairGT encodes node features from $k$-hop, while preserving crucial sensitive features for each node.
This comprehensive approaches not only enhance the graph information encoding but also ensure the independence of sensitive feature, thereby contributing to a fairness-aware training process.Our main contributions can be summarized as follows:
\begin{itemize}
  \item To our best knowledge, this work presents the first attempt to address the fairness issue in GTs.
  We propose FairGT, a novel fairness-aware Graph Transformer, which strategically maintains the independence of sensitive features in the training process.
  \item We introduce an innovative eigenvector selection mechanism into structural topology encoding, followed by a fairness-aware node feature encoding.
  These encodings are designed to ensure the independence of sensitive attributes during the training process of Transformer.
  \item By investigating the correlations between graph information encoding and sensitive features, we provide a theoretical analysis to illustrate the effectiveness of FairGT.
  \item To validate the effectiveness of FairGT, we conduct comprehensive empirical evaluations on five real-world datasets.
  The results showcase the superior performance of FairGT when compared to existing state-of-the-art graph Transformers, GNNs, and fairness-aware GNNs.
\end{itemize}

\section{Related Work}
\subsection{Graph Transformers}
With the advancement of Transformer architectures, it has been observed that leveraging the global receptive field of Transformers on graphs yields effectiveness \cite{ying2021do}.
For instance, GraphTrans \cite{wu2021representing} employs Transformer-based self-attention to acquire long-range pairwise relationships, incorporating a novel readout mechanism to derive a global graph embedding.
SAN \cite{kreuzer2021rethinking} utilizes Laplacian positional encodings for nodes and integrated two attention mechanisms: one for virtual fully-connected graphs and another for actual graph edges.
However, the above methods are mostly designed based on Laplacian positional encodings (that is to calculate the whole Laplacian matrix), which requires $O(n^3)$ complexity.
NAGphormer \cite{chen2023nagphormer} encodes structural topology using eigenvector selection from the Laplacian matrix, and conceptualize each node as a sequence composed of tokens.
This module aggregates neighborhood features across multiple hops, generating diverse representations and forming a sequence of token vectors as input for each node.
Gapformer \cite{liu2023gapformer} converts large-scale graph nodes into a reduced set via local or global pooling, enabling attention computation exclusively with these pooling nodes. This mitigates the impact of irrelevant nodes while maintaining long-range information and reducing computational complexity to linearity.
Nevertheless, it is imperative to acknowledge that prevailing GTs typically lack explicit considerations for algorithmic fairness. Results obtained from the application of these methods to real-world datasets often expose discernible fairness issues.
\begin{figure*}[htbp]
	\centering
	\includegraphics[width=0.9\textwidth]{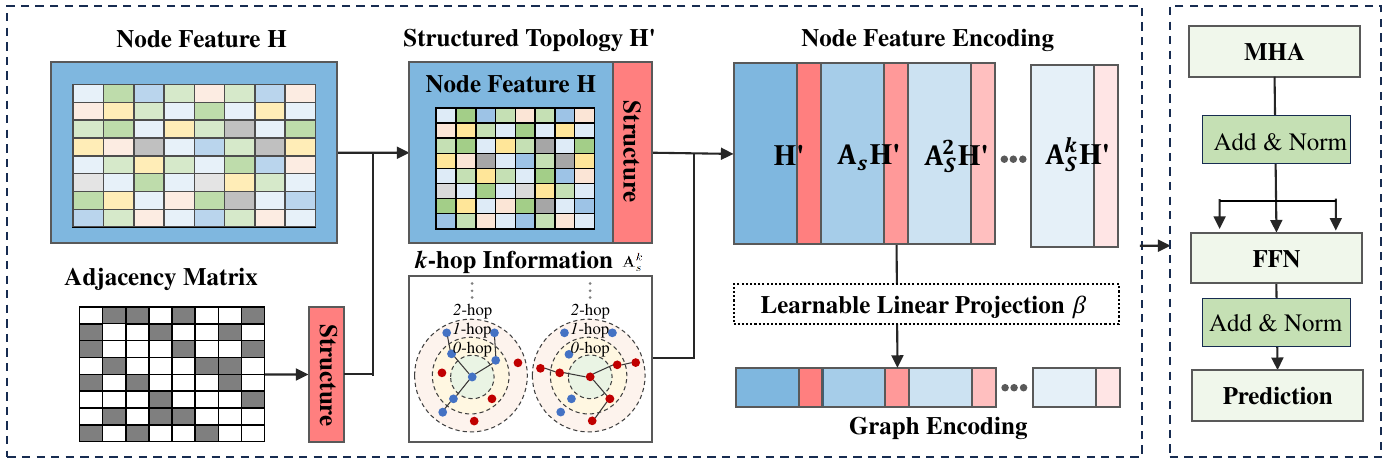}
  \vspace{-0.5em}
  \caption{The illustration of FairGT.}
  \label{fig:architecture}
  \vspace{-1em}
\end{figure*}

\subsection{Fairness-aware GNNs}
Fairness-aware GNNs have garnered significant attention \cite{wang2022fairness,cheng2024data}, which primarily revolve around two methodologies: pre-processing and in-processing.
Some existing fairness-aware methods adopt pre-processing techniques.
For example, Graphair \cite{ling2023learning} automatically identifies fairness-aware augmentations within input graphs based on the structural topology, aiming to circumvent sensitive features while retaining other pertinent information.
However, the unique architecture of GTs, enabling direct node-to-node interactions, is different from traditional structural topology \cite{yin2023lgi}.
The pre-processing techniques may not be directly applicable to balancing input sensitive features of GTs.
In-processing methods focus on making sensitive features independent by altering loss functions or imposing constraints.
For instance, NIFTY \cite{agarwal2021towards} optimizes alignment between predictions derived from perturbed and unperturbed sensitive features.
Similarly, FairGNN \cite{dai2023learning} is an innovative model restricting sensitive features via estimation functions and adversarial debiasing loss. SRGNN \cite{zhang2024learning} presents a fair structural rebalancing algorithm, leveraging gradient constraints to disentangle node representations from sensitive features. However, these constraint-based methods may encounter limitations when applied to GTs owing to the unique nature of graph information encoding.

\section{Preliminaries}

\subsection{Notations}
We first introduce the major notations used throughout the paper in Table \ref{tab:notations}.
Unless otherwise specified, we denote set with copperplate uppercase letters (e.g., $\mathcal{A}$), matrices with bold uppercase letters (e.g., $\mathbf{A}$), and vectors with bold lower-case letters (e.g., $\mathbf{x}$). 
We denote a graph as $\mathcal{G} = \{\mathcal{V}, \mathbf{A}, \mathbf{H}\}$, where $\mathcal{V}$ is the set of $n$ nodes in the graph, $\mathbf{A} \in \mathbb{R}^{n \times n}$ is the adjacency matrix, and $\mathbf{H} \in \mathbb{R}^{n \times d}$ is the node feature matrix.

We use rules similar to NumbPy in Python for matrix and vector indexing.
$\mathbf{A}[i,j]$ represents the entry of matrix $\mathbf{A}$ at the $i$-th row and the $j$-th column.
$\mathbf{A}[i,:]$ and $\mathbf{A}[:,j]$ represent the $i$-th row and the $j$-th column of matrix $\mathbf{A}$, respectively.
One column of $\mathbf{H}$ is the sensitive feature of nodes, we denote it as $\mathbf{H}[:,s]$.
%We denote $v_{s}$ as the sensitive feature of node $v$.
For the binary sensitive feature, $s \in \{0, 1\}$.

\begin{table}[H]
  \vspace{-1em}
  \centering
  \small
  \begin{tabular}{ll}
  \toprule
    \textbf{Notations}            &\textbf{Definitions and Descriptions} \\
  \midrule
    $\mathcal{G}$               & graph set\\
    $\mathcal{V}$               & node set\\
    $\mathbf{A}$                & adjacency matrix\\
    $\mathbf{H}$                & node feature matrix\\
    $n$                         & number of nodes\\
    $d$                         & dimension of feature vector\\
    $k$                         & number of hops\\
    $t$                         & number of eigenvectors\\
    $l$                         & number of layers\\
    $s$                         & sensitive feature\\
  \bottomrule
  \end{tabular}
  \caption{Notations.}
  \vspace{-0.5em}
  \label{tab:notations}
  \vspace{-1em}
\end{table}

\subsection{Fairness Evaluation Metrics}
We present one definition of fairness for the binary label $y \in \{0,1\}$ and the sensitive feature $s \in \{0,1\}$, $\hat{y} \in \{0,1\}$ denotes the class label of the prediction.

\textbf{Statistical Parity} \cite{dwork2012fairness} (i.e., Demographic Parity and Independence). Statistical parity requires the predictions to be independent of the sensitive features $s$. It could be formally written as:
\begin{equation}
    \mathbb{P}(\hat{y}|s=0)=\mathbb{P}(\hat{y}|s=1).
\label{equ:SP}
\end{equation}

When both the predicted labels and sensitive features are binary, to quantify the extent of statistical parity, the $\Delta_{\text{SP}}$ is defined as follows:
\begin{equation}
  \Delta_{\text{SP}}=|\mathbb{P}(\hat{y}=1|s=0)-\mathbb{P}(\hat{y}=1|s=1)|.
\label{equ:delta_SP}
\end{equation}

The $\Delta_{\text{SP}}$ measures the acceptance rate difference between the two sensitive subgroups.

\subsection{Transformer}
The Transformer architecture consists of a composition of Transformer layers.
Each Transformer layer has two parts: a self-attention module and a position-wise feed-forward network (FFN).
Let $\mathbf{X} = [h_1^\top, ... , h_i^\top]^\top \in \mathbb{R}^{i \times d_m}$ denotes the input of self-attention module where $d_m$ is the hidden dimension and $h_j \in \mathbb{R}^{1 times d_m}$ is the hidden representation at position $j$.
The input $\mathbf{X}$ is projected by three matrices $W_Q \in \mathbb{R}^{d_m \times d_K}$, $W_K \in \mathbb{R}^{d_m \times d_K}$ and $W_V \in \mathbb{R}^{d_m \times d_V}$ to the corresponding representations $Q$, $K$, $V$.
The self-attention is then calculated as:
\begin{equation}
  \label{equ:Transformer 1}
  Q = \mathbf{X}W_Q, K = \mathbf{X}W_K, V = \mathbf{X}W_V,
\end{equation}
\begin{equation}
  \label{equ:Transformer 2}
  \text{Attn}(\mathbf{X})=\text{softmax}(\frac{QK^\top}{\sqrt{d_K}})V.
\end{equation}
For simplicity of illustration, we consider the self-attention and assume $d_K=d_V=d$.
The extension to the multi-head attention is straightforward, and we omit bias terms for simplicity.

\section{The Design of FairGT}
In this section, we first present our theoretical findings that underpin FairGT, followed by the design of structural topology encoding and node feature encoding. 
The architecture of FairGT (see Figure \ref{fig:architecture}) is then introduced, as well as its complexity analysis.
\subsection{Theoretical Findings Underpinning FairGT}
For a fairness-aware structural topology encoding, we analyze the similarity between the distribution of original sensitive features and the distribution of $k$-hop neighbor sensitive features.

\textbf{Lemma 1}
\textit{The similarity between the distribution of original sensitive features and the distribution of $k$-hop neighbor sensitive features exhibits a pronounced correlation with the eigenvector corresponding to the largest magnitude eigenvalue of the adjacency matrix. Concurrently, the correlation with other eigenvectors diminishes exponentially.}

\textit{Proof.}
Assume $\mathbf{A} \in \mathbb{R}^{n \times n}$ is an adjacency matrix with real-valued entries.
The $k$-hop neighbor feature matrix is $\mathbf{A}^k \mathbf{H}$, and the $k$-hop neighbor sensitive features are $\mathbf{A}^k \mathbf{H}[:,s]$.
$|\lambda_1| > |\lambda_2| \geq ......  \geq |\lambda_n|$ are $n$ real eigenvalues, and $\mathbf{p}_i$ ($i \in \{1, 2, ......, n\}$) are corresponding eigenvectors.
We assume $\alpha_i = \mathbf{H}[:,s]^ \top \mathbf{p}_i$.
We use cosine similarity to measure the similarity.
Thus, the following equation establishes:
\begin{equation}
  \nonumber
  cos(<\mathbf{p}_i, \mathbf{H}[:,s]>)=\frac{\alpha_i}{\sqrt{\sum^n_{j=1}\alpha_j^2}}.
\end{equation}

Then,
\begin{equation}
  \nonumber
  \begin{aligned}
  &cos(<\mathbf{A}^k \mathbf{H}[:,s], \mathbf{H}[:,s]>) \\
  =&\frac{\alpha_1^2 + \sum^n_{i=2}\alpha_i^2 (\frac{\lambda_i}{\lambda_1})^l}{\sqrt{\alpha_1^2 + \sum^n_{i=2}\alpha_i^2 (\frac{\lambda_i}{\lambda_1})^{2l}}\sqrt{\sum^n_{i=1}\alpha_i^2}}.
  \end{aligned}
  \label{equ:eigenvector selection}
\end{equation}

Thus, the correlation between $\mathbf{p}_i$ and $cos(<\mathbf{A}^k \mathbf{H}[:,s], \mathbf{H}[:,s]>)$ is proportional to $(\frac{\lambda_i}{\lambda_1})^l$.
Because of $|\lambda_1| > |\lambda_i|$, the correlation decays exponentially.
Specifically, when $k \rightarrow \infty$, we have:
\begin{equation}
  \nonumber
  \lim_{k \to \infty} cos(<\mathbf{A}^k \mathbf{H}[:,s], \mathbf{H}[:,s]>) = cos(<\mathbf{p}_1, \mathbf{H}[:,s]>).
\end{equation}

%Details of this proof are shown in Appendix A (Supplementary Material).\\
\rightline{$\square$}

To ensure the node feature encoding to be independent with sensitive features, we divide the nodes of the origin graph $\mathcal{G}$ into two subgraphs $\mathcal{G}_0$, $\mathcal{G}_1$.
Specifically, $\mathcal{G}_0 = \{v \in \mathcal{G} | v_{sen} = 0\}$ and $\mathcal{G}_1 = \{v \in \mathcal{G} | v_{sen} = 1\}$.
Both $\mathcal{G}_0$ and $\mathcal{G}_1$ are complete subgraphs.
We propose a sensitive feature complete graph, $\mathcal{G}_{s} = \mathcal{G}_0 \cup \mathcal{G}_1$.
Given the adjacency matrix $\mathbf{A}_{s}$ of $\mathcal{G}_{s}$ and $\mathbf{H}$, multiplying $\mathbf{A}_{s}$ with $\mathbf{H}$ aggregates $k$-hop sensitive information.
Applying this multiplication consecutively allows us to propagate information at larger distances.
For example, we can access 2-hop sensitive information by $\mathbf{A}_{s} (\mathbf{A}_{s}\mathbf{H})$.
Thereafter, the $k$-hop sensitive matrix can be described as:
\begin{equation}
  \label{equ:sensitive matrix}
  \mathbf{H}^{(k)} = \mathbf{A}_{s}^k\mathbf{H}.
\end{equation}

\textbf{Lemma 2}
\textit{The sensitive feature distribution of $\mathbf{H}^{(k)}$ is the same as $\mathbf{H}$:}
\begin{equation}
  \mathbf{H}^{(k)}[i, s] = q^k\mathbf{H}[i, s],
\end{equation}
\textit{where $i \in \{1,2,..., n\}$, q denotes the number of nodes whose sensitive feature is 1.}

\textit{Proof}. We prove this lemma by mathematical induction.
We firstly prove the based case ($k$=1).
Then, we give the inductive hypothesis ($k$=$r$), and confirm the inductive step ($k=r+1$).
Combining the foundational step, the inductive hypothesis, and the inductive step, we show that this mathematical statement holds for all positive integer $k$.
\rightline{$\square$}

The node feature encodings of FairGT effectively aggregate information from $k$-hop nodes while upholding independence with sensitive features.

\subsection{Structural Topology Encoding}
The structural information of nodes is a crucial feature for graph mining tasks.
Spectral graph theory illustrates that the algebraic connectivity and spectral radius, representing the lowest and largest non-zero eigenvalues respectively, are intricately linked to the geometric properties of the graph \cite{kreuzer2021rethinking}.

Furthermore, inspired by \textbf{Lemma 1}, we adopt the eigenvectors selection of adjacency matrix of the graph for capturing the structural information of nodes.
Specifically, we select the eigenvectors corresponding to the $t$ largest magnitude eigenvalues to construct the structure matrix $\mathbf{B} \in \mathbb{R}^ {n \times t}$.
Then we combine the original feature matrix $\mathbf{H}$ with the structure matrix $\mathbf{B}$ to preserve both node features and structural information:
\begin{equation}
  \label{equ:structural information}
  \mathbf{H}' = \mathbf{H} || \mathbf{B},
\end{equation}
where $||$ indicates the concatenation operator and $\mathbf{H}' \in \mathbb{R} ^{n \times (d+t)}$ denotes the fused feature matrix.
In this structural topology, the similarity between the distribution of sensitive features and the sensitive features of $k$-hop neighbors is high, contributing to a fairer graph information encoding.

\subsection{Node Feature Encoding}
In addition to the structural topology of graph, node features also contain valuable information in a graph.
To address fairness issues in GTs, we introduce a fairness-aware node feature encoding.
This encoding method considers the $k$-hop information independent with sensitive features.

Specifically, for node $v$, let $\mathcal{V}^{(k)}_s = \{v^{(k)} \in \mathcal{V} | v^{(k)}_{s}=v^{(k-1)}_{s}, v^{(k)} \neq v^{(k-1)}\}$ be its $k$-hop information set.
We define $\mathcal{V}^{(0)}_s = {v}$, i.e., the $0$-hop is the node itself.
In node feature encoding, we transform the $k$-hop information set $\mathcal{V}^{(k)}_s$ into a feature embedding $\mathbf{H'}^{(k)}_v$ with an operator $\Phi$.
The operator $\Phi$ serves as an aggregation operator that aggregates $k$-hop information from same sensitive feature neighbors.
In this way, the $k$-hop representation of node $v$ can be expressed as:
\begin{equation}
  \label{equ:k-hop}
  \mathbf{H'}^{(k)}_v=\Phi(v,\mathcal{V}^{(1)}_s,..., \mathcal{V}^{(k)}_s).
\end{equation}

By using Equation \eqref{equ:k-hop}, we can calculate the information for variable hops of a certain node and further construct the corresponding sequence, i.e., $\mathbf{S}_v=(\mathbf{H'}^{(0)}_v, \mathbf{H'}^{(1)}_v, ... , \mathbf{H'}^{(k)}_v)$.
Assume $\mathbf{H'}^{(k)}_v$ is a $d$-dimensional vector, the sequence of all nodes in graph $\mathcal{G}$ will constitute a tensor $\mathcal{H} \in \mathbb{R}^{n \times (k+1) \times d}$.
To better illustrate the implementation of node feature encoding, we decompose $\mathbf{H'}$ to a sequence $\mathbf{S} = (\mathbf{H'}^{(0)}, \mathbf{H'}^{(1)}, ... , \mathbf{H'}^{(k)})$, where $\mathbf{H'}^{(k)} \in \mathbb{R}^{n \times d}$ can be seen as the $k$-hop feature matrix.
Here we define $\mathbf{H'}^{(0)}$ as the original feature matrix.
In the experiments, we create a sensitive feature complete graph $\mathcal{G}_{s}$ (details are illustrated in Section 4.1).
Given the adjacency matrix $\textbf{A}_{s}$ of $\mathcal{G}_{s}$.
The $k$-hop feature matrix is defined as following equation:
\begin{equation}
  \mathbf{H'}^{(k)} = \mathbf{A}_{s}^k\mathbf{H'}.
\end{equation}

According to \textbf{Lemma 2}, the encoding process of node features ensures the independence of sensitive features.
Representing the $k$-hop information following this encoding proves to be beneficial for capturing hop-wise correlations, which is a crucial aspect in fairness-aware graph transformers.

\subsection{FairGT for Node Classification}
Given an graph, we first concatenate a matrix constructed by eigendecomposition to the adjacency matrix, and encode structure topology of graph following Equation \eqref{equ:structural information}.
Next, we assemble an aggregated neighborhood sequence as $\mathbf{S}_v=(\mathbf{H'}^{(0)}_v, \mathbf{H'}^{(1)}_v, ... , \mathbf{H'}^{(k)}_v)$ by applying node feature encoding.
Then we map $\mathbf{S}_v$ to the hidden dimension $d_h$ of the Transformer with a learnable linear projection:
\begin{equation}
  \label{equ:linear projection}
  \mathbf{T}^{(0)}_v = [\mathbf{H'}^{(0)}_v \beta, \mathbf{H'}^{(1)}_v \beta, ... , \mathbf{H'}^{(k)}_v \beta],
\end{equation}
where $\beta \in \mathbb{R}^{d \times d_h}$ and $\mathbf{T}^{(0)}_v \in \mathbb{R}^ {(k+1) \times d_h}$.

Then, following the projection of the sequence representing structural topology, we proceed to input this transformed sequence into the Transformer architecture \cite{vaswani2017attention}.
The building blocks of the Transformer contain multi-head self-attention (MHA) and position-wise feed-forward network (FFN), wherein LayerNorm(LN) is applied before each block. The FFN consists of two linear layers with a GELU non-linearity:
\begin{equation}
    \begin{aligned}
        &\mathbf{T}'^{(l)}_v = \textbf{MHA}(\textbf{LN}(\mathbf{T}'^{(l-1)}_v)) + \mathbf{T}'^{(l-1)}_v, \\
        &\mathbf{T}^{(l)}_v = \textbf{FFN}(\textbf{LN}(\mathbf{T}'^{(l)}_v)) + \mathbf{T}'^{(l)}_v,
    \end{aligned}
\label{equ:Transformer}
\end{equation}
where $l = 1, 2, ... , L$ implies the $l$-th layer of the Transformer.

In the end, we obtain the prediction by applying weighted summation to the output of the encoder using attention mechanisms.
Through several Transformer layers, the corresponding output $\mathbf{T}^{(l)}_v$ contains the embeddings for all neighborhoods of node $v$.
It requires a weighted summation to aggregate the information of $k$-hop node features into one. Conventional summation methods often overlook the significance of diverse neighborhoods. In addressing this limitation, we employ attention-based summation to learn the importance of various neighborhoods by computing the attention coefficients.

\subsection{Complexity of FairGT}
Existing GTs treat the nodes as independent tokens and construct a single sequence composed of all the node tokens to train the Transformer model, suffering from a quadratic complexity on the number of nodes for the self-attention calculation.
Because FairGT encodes node features as one vector, the time complexity of FairGT is $O(n(k+1)^2d)$ and the space complexity of FairGT is $O(n(k+1)^2 + n(k+1)d + d^2l)$.

Furthermore, eigendecomposition causes cubic complexity in the number of nodes, which captures the node structural information, resulting in a computational cost of $O(n^3)$.
In the eigenvector selection process of FairGT, we donnot need to rank the eigenvector after eigendecomposition.
Within the Arnoldi Package algorithm \cite{richard1998arpack}, the time complexity of eigendecomposition in FairGT is $O(nt^2)$, and the space of complexity is $O(n)$.

\section{Experiments}
\subsection{Datasets}
In this study, node classification serves as the downstream task, employing five distinct real-world datasets: \textbf{NBA}, \textbf{Bail}, \textbf{German}, \textbf{Credit}, and \textbf{Income}. The statistics of five datasets are shown in Table \ref{tab:datasets}.
\begin{itemize}
  \item \textbf{NBA} \cite{dai2021say}: The NBA dataset, sourced from Kaggle, encompasses player statistics from the 2016-2017 season, including additional player information and Twitter relationships.
  It categorizes nationality as U.S. or overseas, focusing on predicting whether a player's salary exceeds the median.
  \item \textbf{Bail} \cite{jordan2015effect}: This dataset represents defendants who got released on bail at the U.S. state courts during 1990-2009, connected by edges based on shared past criminal records and demographics.
  The goal is predicting a defendant's likelihood of committing either a violent or nonviolent crime post-release, with 'race' as the sensitive feature.
  \item \textbf{German} \cite{asuncion2007uci}: German is extracted from the Adult Data Set .
  The dataset is a credit graph which has 1,000 nodes representing clients in a German bank that are connected based on the similarity of their credit accounts.
  The objective is to categorize clients' credit risk as either high or low, with “gender” designated as the sensitive feature.
  \item \textbf{Credit} \cite{yeh2009the}: The Credit dataset uses personal next month.
  Comprising individuals connected based on similarities in spending and payment patterns.
  Age serves as the sensitive feature, while the label feature denotes defaulting on credit card payments.
  \item \textbf{Income} \cite{asuncion2007uci}: Income is extracted from the Adult Data Set.
  Each node represents an individual, with connections established based on criteria similar to \cite{agarwal2021towards}.
  The sensitive feature in this dataset is race, and the task involves classifying whether an individual's salary exceeds $50,000$ annually.
\end{itemize}

\begin{table}[H]
  \centering
  \small
  \tabcolsep=0.1cm
  \begin{tabular}{ccccc}
  \toprule
  \textbf{Dataset}    & \textbf{\# Nodes} & \textbf{\# Edges} & \textbf{Sensitive feature} & \textbf{Label}\\
    \midrule
      \textbf{NBA}    & $403$     & $16,570$  & Nationality & Salary \\
      \textbf{German} & $1,000$   & $22,242$  & Gender      & Good/Bad Customer \\
      \textbf{Bail}   & $18,876$  & $321,308$ & Race        & Recidivism \\
      \textbf{Credit} & $30,000$  & $137,377$ & Age         & Default\\
      \textbf{Income} & $14,821$  & $100,483$ & Race        & Income\\
    \bottomrule
  \end{tabular}
  \caption{The statistics of the five real-world datasets. }
  \vspace{-1em}
  \label{tab:datasets}
\end{table}
\begin{table*}
  \centering
  \small
  \tabcolsep=0.3cm
  \begin{tabular}{lcccccccccc}
    \toprule
    \multirow{2}[2]{*}{\textbf{Methods}} & \multicolumn{2}{c}{\textbf{NBA}} & \multicolumn{2}{c}{\textbf{Bail}} & \multicolumn{2}{c}{\textbf{German}} & \multicolumn{2}{c}{\textbf{Credit}} & \multicolumn{2}{c}{\textbf{Income}} \\
    \cmidrule{2-11}
    & ACC $\uparrow$ & $\Delta_{\text{SP}}$ $\downarrow$ & ACC $\uparrow$ & $\Delta_{\text{SP}}$ $\downarrow$ & ACC $\uparrow$ & $\Delta_{\text{SP}}$ $\downarrow$ & ACC $\uparrow$ & $\Delta_{\text{SP}}$ $\downarrow$ & ACC $\uparrow$ & $\Delta_{\text{SP}}$ $\downarrow$\\
    \midrule
    \textbf{GCN} & $71.70$ & $9.18$ & $84.56$ & $7.35$ & $73.44$ & $35.17$ & $73.87$ & $12.86$ & $73.87$ & $25.93$\\
    \textbf{GCNII} & $72.68$ & $14.17$ & $92.39$ & $5.67$ & $71.60$ & $6.81$ & $73.95$ & $16.85$ & $76.20$ & $16.20$\\
    \textbf{GAT} & $72.45$ & $11.54$ & $93.24$ & $5.44$ & $72.80$ & $12.52$ & $68.29$ & $9.74$ & $69.14$ & $12.46$ \\
    \textbf{Specformer} & $73.42$ & $11.59$ & $95.46$ & $6.32$ & $72.40$ & $6.90$ & $74.06$ & $8.37$ & $80.06$ & $7.26$\\
    \midrule
    \textbf{FairGNN} & $69.72$ & $1.32$ & $82.94$ & $6.90$ & $69.68$ & $3.49$ & $68.29$ & $9.74$ & $69.14$ & $12.46$\\
    \textbf{NIFTY} & $70.16$ & $3.26$ & $83.43$ & $4.75$ & $69.92$ & $5.73$ & $66.81$ & $13.59$ & $70.87$ & $24.43$\\
    \textbf{BIND} & $72.14$ & $4.37$ & $89.72$ & $7.77$ & $71.60$ & $3.46$ & $68.64$ & $11.65$ & $71.76$ & $14.75$\\
    \textbf{Graphair} & $70.99$ & $2.54$ & $84.76$ & $4.98$ & $70.40$ & $6.39$ & $67.68$ & $8.99$ & $71.50$ & $10.68$\\
    \midrule
    \textbf{GraphTrans} & $73.81$ & $9.01$ & $93.56$ & $7.03$ & $74.40$ & $8.27$ & $71.35$ & $8.59$ & $79.57$ & $8.89$\\
    \textbf{SAN} & $73.41$ & $29.02$ & $74.91$ & $6.97$ & $73.20$ & $9.77$ & $70.01$ & $8.39$ & $79.44$ & $7.41$\\
    \textbf{NAGphormer} & $72.15$ & $16.74$ & $93.28$ & $7.03$ & $75.20$ & $8.27$ & $77.81$ & $8.29$ & $80.17$ & $7.49$\\
    \midrule
    \textbf{FairGT} & \textbf{74.68} & \textbf{0.38} & \textbf{95.68} & \textbf{0.58} & \textbf{76.00} & \textbf{0.26} & \textbf{77.85} & \textbf{1.89} & \textbf{81.30} & \textbf{2.66}\\
    \bottomrule
  \end{tabular}
  \caption{Comparison of performance (accuracy) and fairness ($\Delta_{\text{SP}}$) in percentage (\%). $\uparrow$ denotes the larger, the better; $\downarrow$ denotes the opposite. The best results are bold-faced.}
  \vspace{-1em}
  \label{tab:result}
\end{table*}

Considering some datasets includes more than two classes of ground truth labels, we save the class of labels $0$ and $1$ and change the class of labels more than $1$ to $1$.
Then, we randomly select $25$\% nodes as the validation set and $25$\% as the test set, ensuring a balanced ratio of ground truth labels.
On \textbf{NBA}, we randomly select either $50$\% nodes or $50$ nodes in each class of ground truth labels, depending on which is a smaller number.
On \textbf{Bail}, we change the number of nodes from $50$ to $100$.
On \textbf{German}, \textbf{Credit}, and \textbf{Income}, we change $50$ to $500$.
Such a splitting strategy is also followed by fairness-aware baselines \cite{agarwal2021towards,dai2023learning,dong2023interpreting}.

\subsection{Baselines}
To ensure diversity, we use three classes of baselines: GNNs, GTs, and fairness-aware GNNs.
\begin{itemize}
  \item \textbf{GNNs}: We use two GNNs: GCN \cite{kipf2017semi} and GCNII \cite{chen2020simple}.
  We also used two GNNs based on attention mechanism: GAT \cite{velickovic2018graph} and Specformer \cite{bo2023specformer}.
  \item \textbf{Fairness-aware GNNs}: We use four fairness-aware GNNS: FairGNN \cite{dai2023learning}, NIFTY \cite{agarwal2021towards}, BIND \cite{dong2023interpreting}, and Graphair \cite{ling2023learning}.
  \item \textbf{GTs}: We use three representative GTs: GraphTrans \cite{dwivedi2020generalization}, SAN \cite{kreuzer2021rethinking}, and NAGphormer \cite{chen2023nagphormer}.
\end{itemize}

\subsection{Comparison Results}
Table \ref{tab:result} provides a comprehensive overview of the fairness evaluation metrics, concerning our proposed FairGT method and various baseline models across five real-world datasets.
We report the overall Accuracy and $\Delta_{\text{SP}}$ (average results of five-fold cross-validation) respectively.
This table shows that FairGT consistently demonstrates outstanding fairness, proving the effectiveness of our method in the fairness-aware node classification.
FairGT not only enhances fairness compared to GTs but also outperforms existing fairness-aware GNN methods in terms of fairness results. This underscores its efficacy in addressing fairness-related concerns within GTs. Furthermore, FairGT demonstrates improved performance in node classification across five real-world datasets.

\subsection{Training Cost Comparison}
FairGT, utilizing a subset of adjacency matrix eigenvectors and employing vectors as tokens, may exhibit efficiency compared to GTs.
To validate the efficiency of FairGT, we compare its training time with GT baselines. For a fair comparison, we standardize key parameters across all methods, setting the number of hidden dimensions to 128, the number of layers to 1, the number of heads to 1, and the number of epochs to 500.

\begin{table}[H]
  \centering
  \small
  \tabcolsep=0.15cm
  \begin{tabular}{lcccc}
    \toprule
    \textbf{Dataset} & \textbf{FairGT} & \textbf{GraphTrans} & \textbf{SAN} & \textbf{NAGphormer} \\
    \midrule
    \textbf{NBA}      & $17.87$   & $24.06$   & $36.98$   & $20.21$   \\
     \textbf{Bail}    & $116.95$  & $168.04$  & $203.01$  & $137.06$  \\
     \textbf{German}  & $19.82$   & $33.91$   & $39.56$   & $21.07$   \\
     \textbf{Credit}  & $217.35$  & $241.94$  & $252.22$  & $186.42$  \\
     \textbf{Income}  & $124.99$  & $118.29$  & $140.55$  & $119.90$  \\
    \bottomrule
  \end{tabular}
  \caption{Runtime (s) of GTs over five real-world datasets.}
  \label{tab:cost time}
  \vspace{-1em}
\end{table}

The results are summarized in Table \ref{tab:cost time}, which demonstrate that our method attains superior performance and fairness in node classification without significantly escalating computational complexity.

\begin{figure*}[htbp]
	\centering
	\subfigure {
		\begin{minipage}[b]{0.185\textwidth}
			\centering
			\includegraphics[width=1\textwidth]{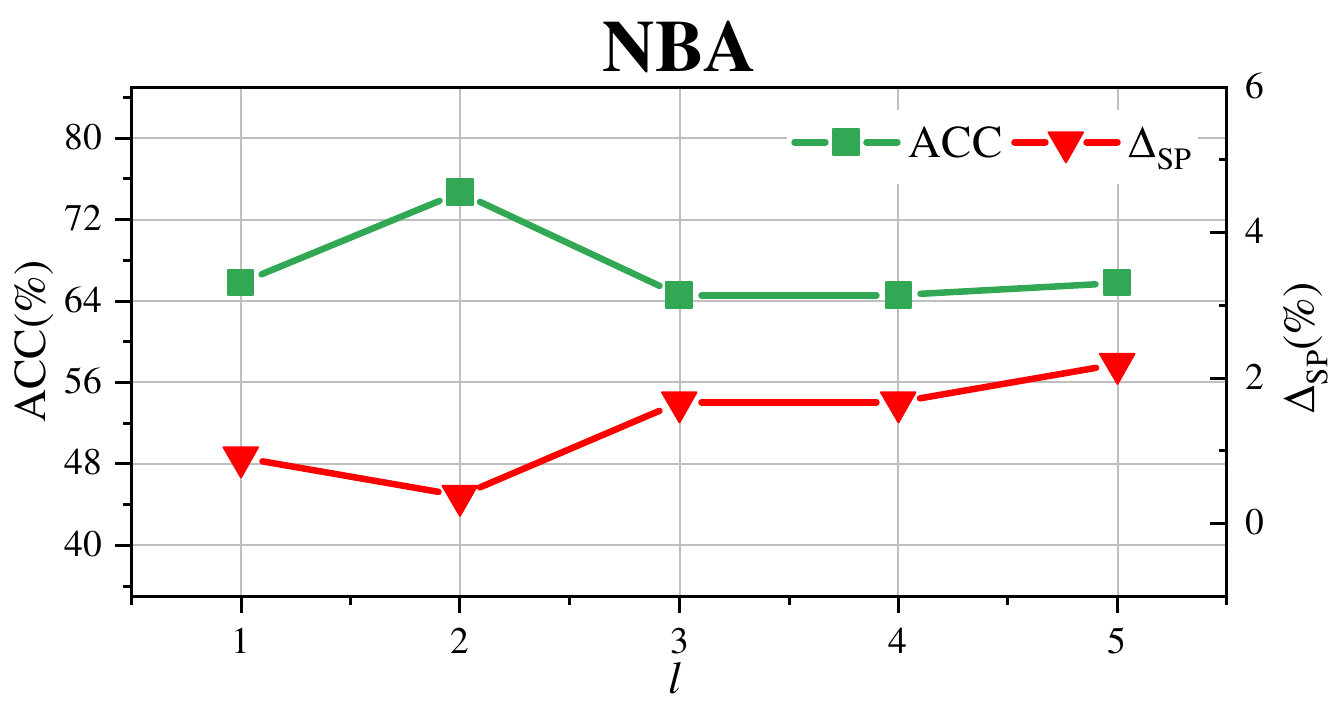}
		\end{minipage}
	}
	\subfigure {
		\begin{minipage}[b]{0.185\textwidth}
			\centering
			\includegraphics[width=1\textwidth]{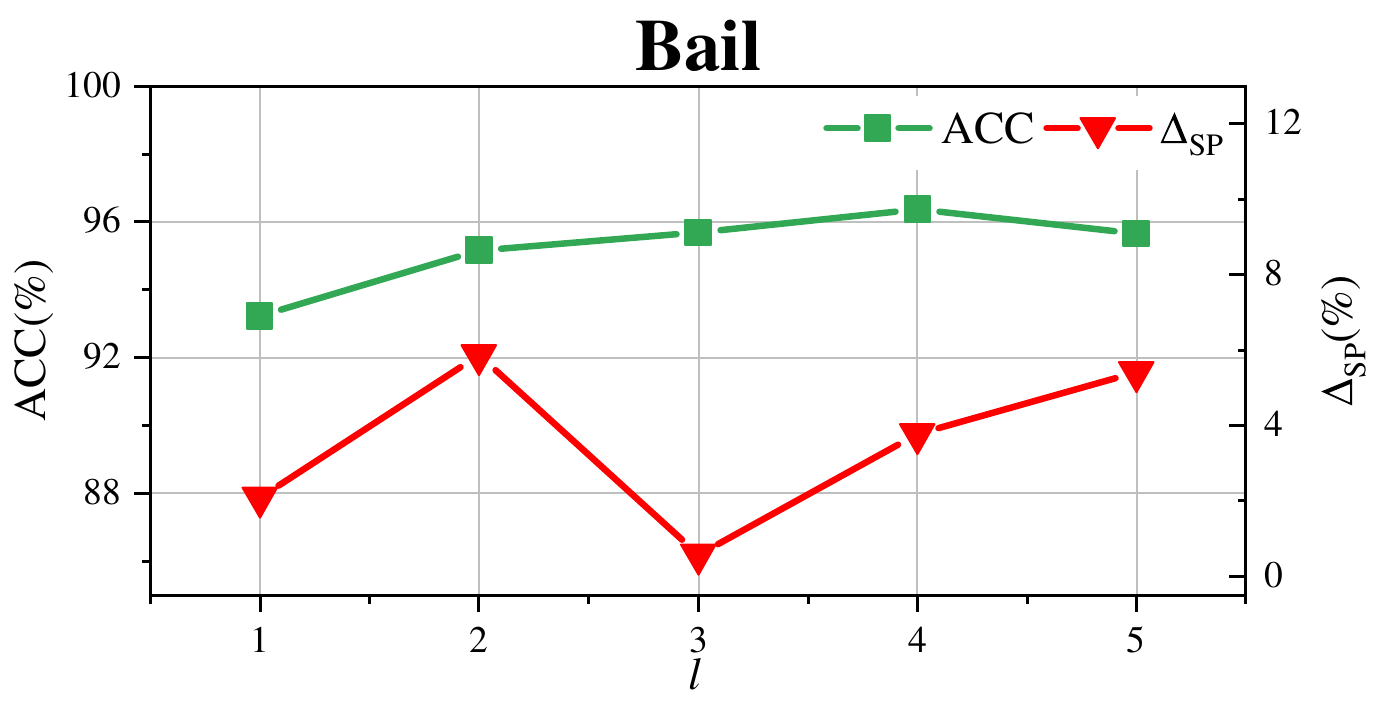}
		\end{minipage}
	}
	\subfigure {
   	\begin{minipage}[b]{0.185\textwidth}
   		\centering
   		\includegraphics[width=1\textwidth]{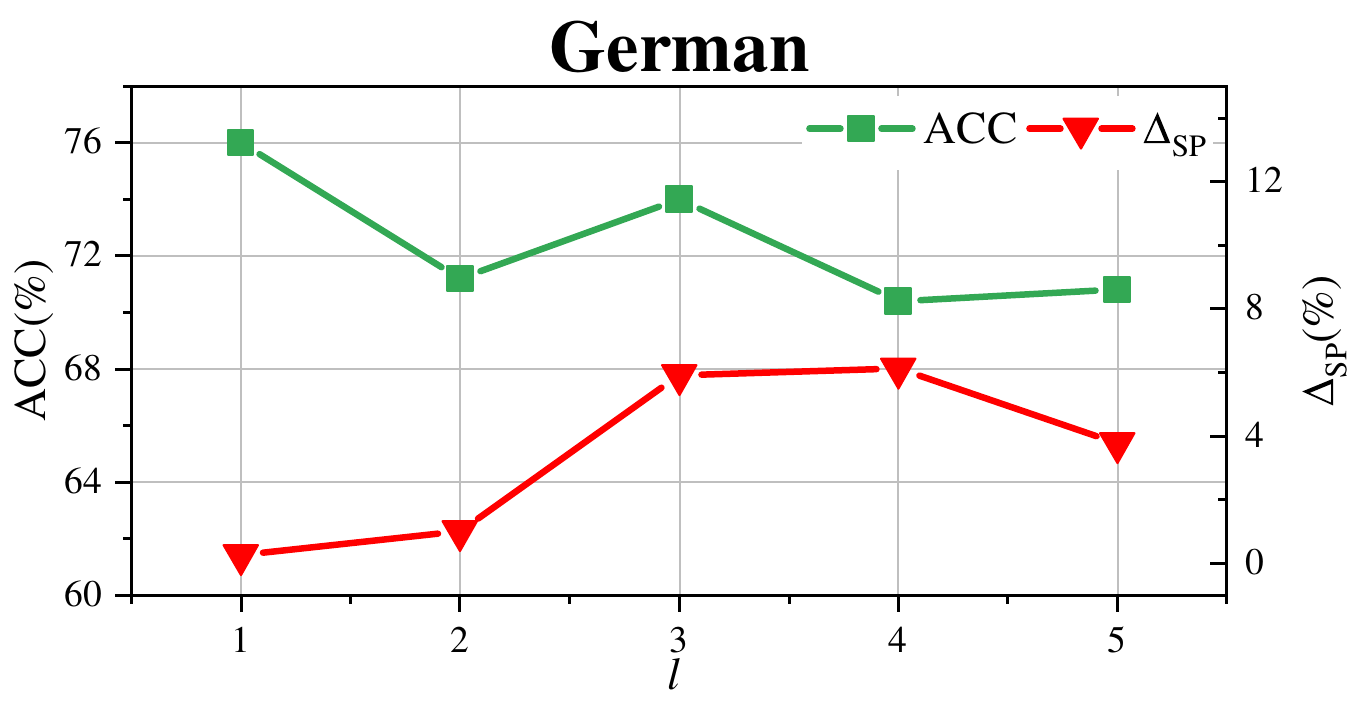}
  	\end{minipage}
  }
  \subfigure {
    \begin{minipage}[b]{0.185\textwidth}
      \centering
      \includegraphics[width=1\textwidth]{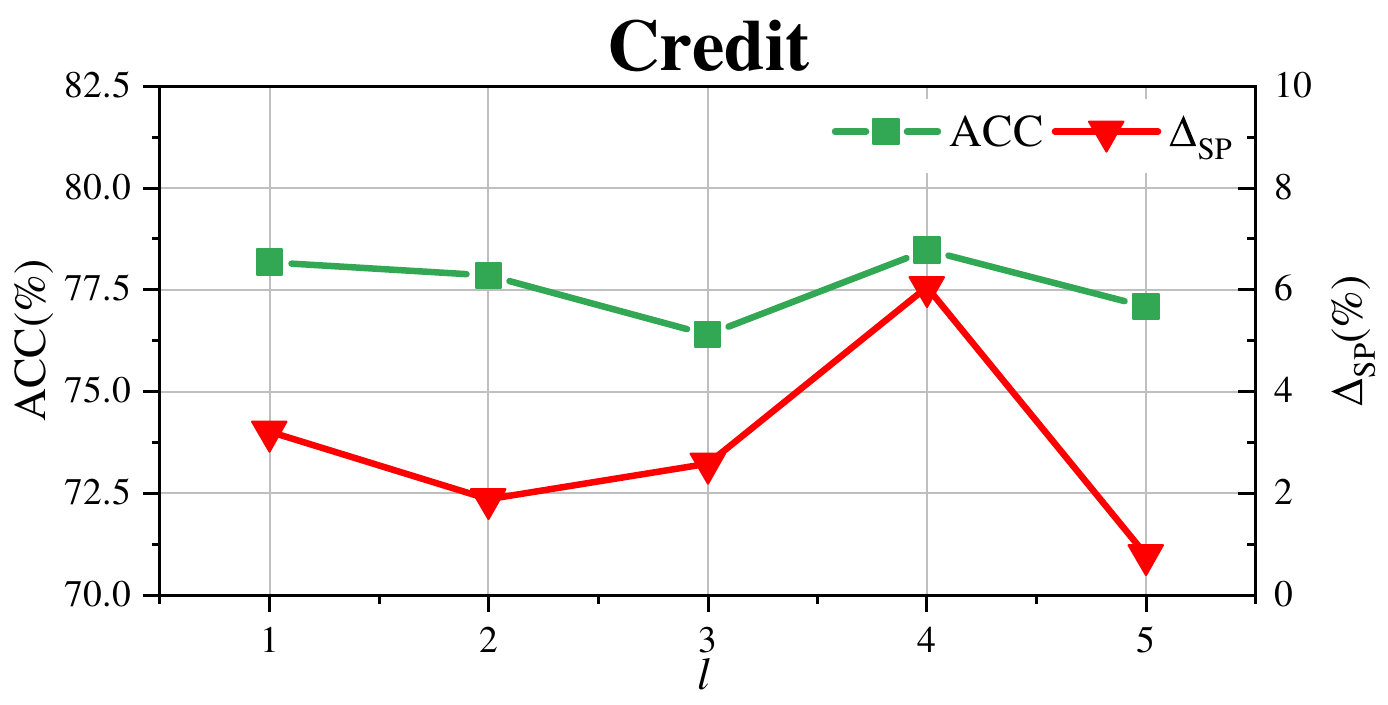}
    \end{minipage}
  }
  \subfigure {
    \begin{minipage}[b]{0.185\textwidth}
      \centering
      \includegraphics[width=1\textwidth]{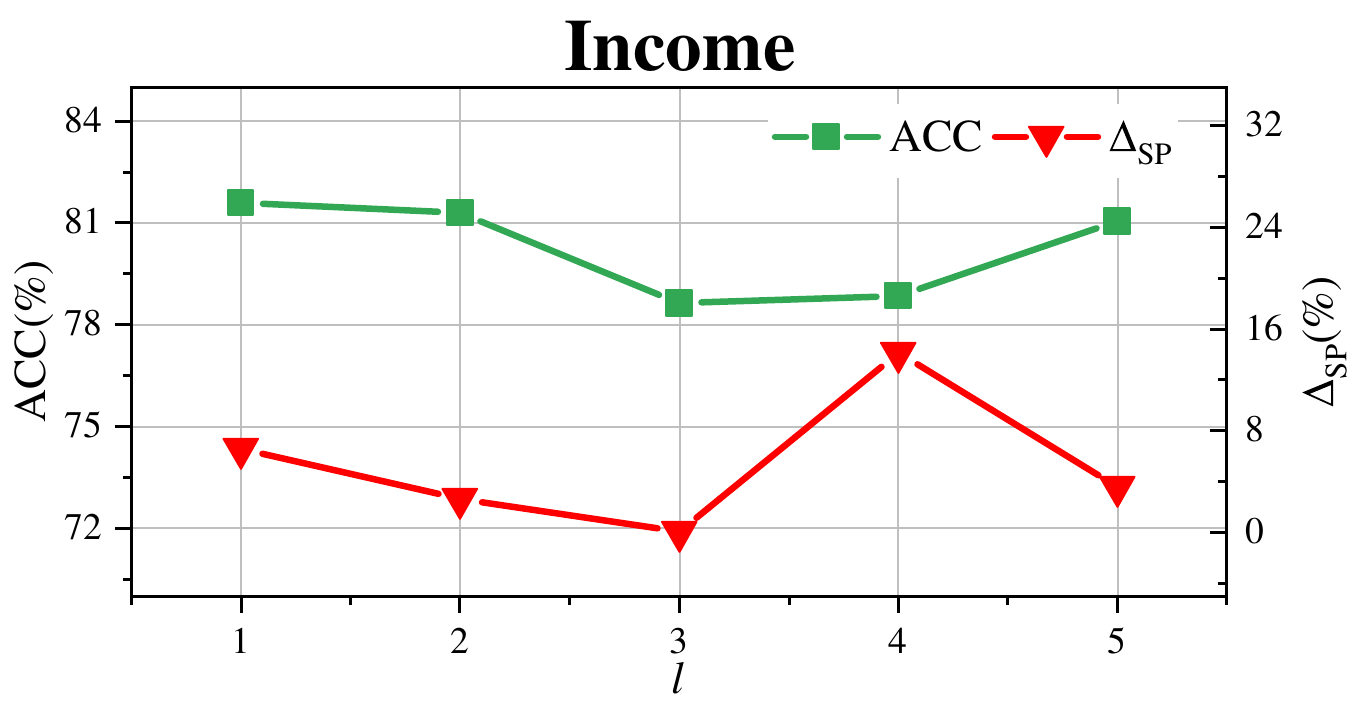}
    \end{minipage}
  }
  \vspace{-1em}
  \caption{The accuracy and $\Delta_{\text{SP}}$ of FairGT w.r.t. different parameter $l$ values.}
  \label{fig:parameter t}
  \vspace{-1em}
\end{figure*}

\begin{figure*}[htbp]
	\centering
	\subfigure {
		\begin{minipage}[b]{0.185\textwidth}
			\centering
			\includegraphics[width=1\textwidth]{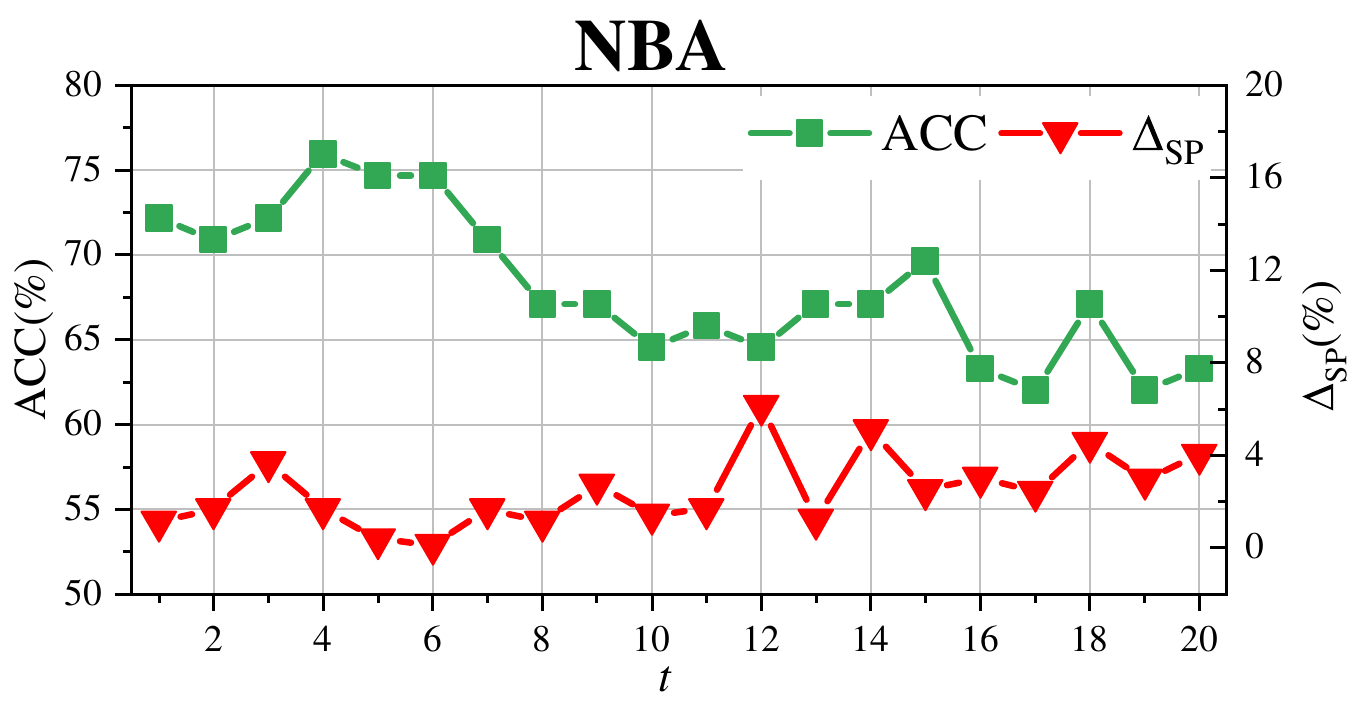}
		\end{minipage}
	}
	\subfigure {
		\begin{minipage}[b]{0.185\textwidth}
			\centering
			\includegraphics[width=1\textwidth]{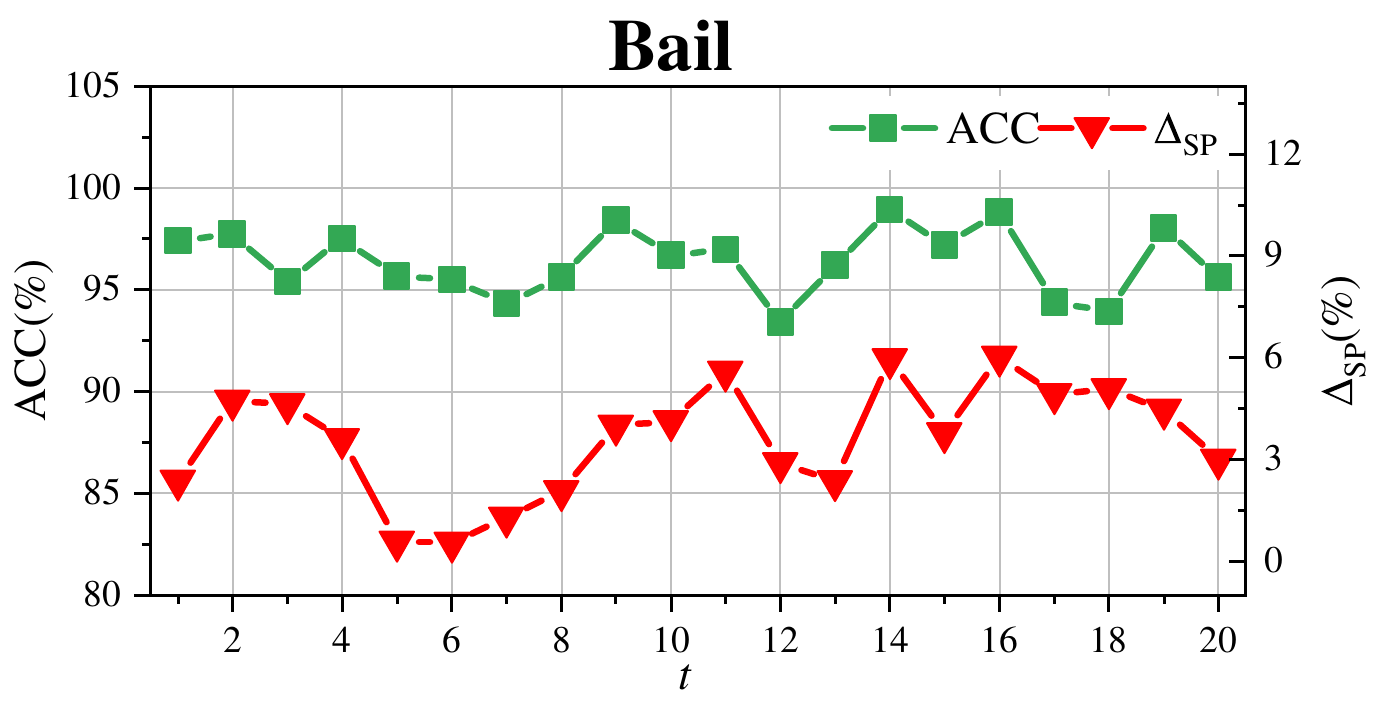}
		\end{minipage}
	}
	\subfigure {
   	\begin{minipage}[b]{0.185\textwidth}
   		\centering
   		\includegraphics[width=1\textwidth]{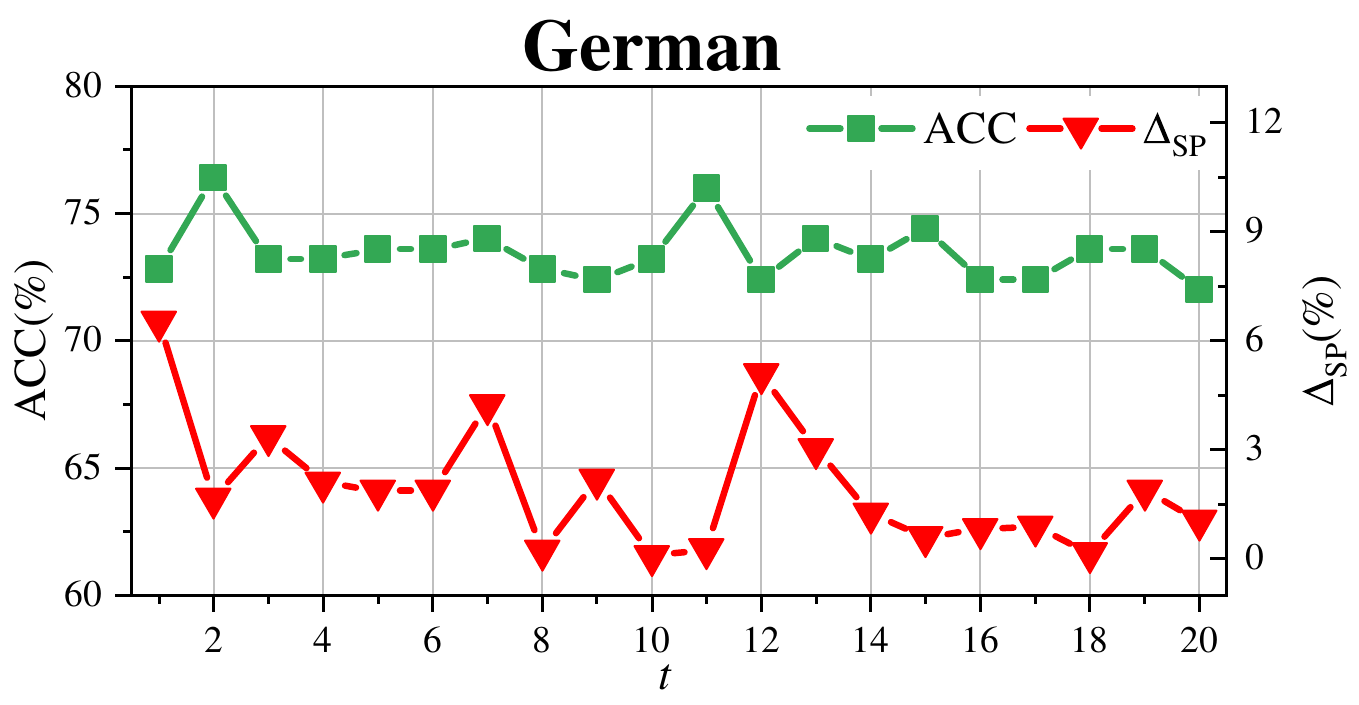}
  	\end{minipage}
  }
  \subfigure {
    \begin{minipage}[b]{0.185\textwidth}
      \centering
      \includegraphics[width=1\textwidth]{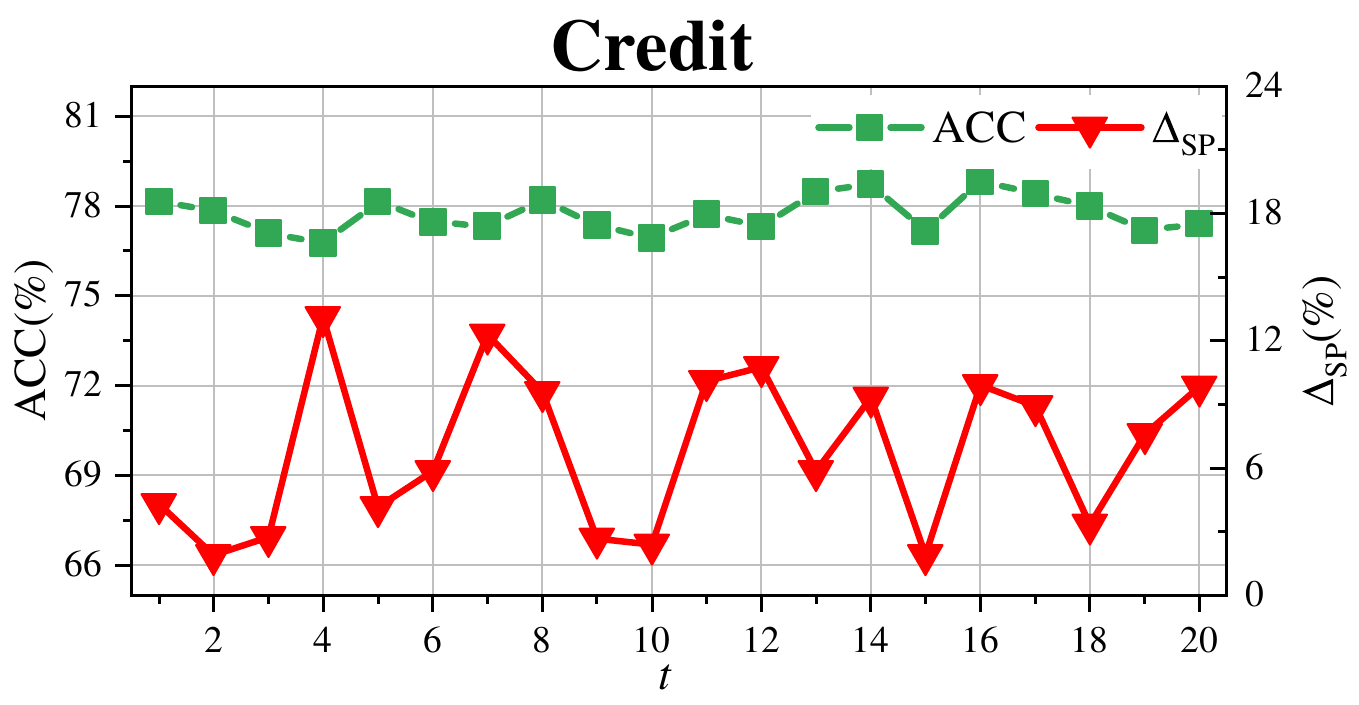}
    \end{minipage}
  }
  \subfigure {
    \begin{minipage}[b]{0.185\textwidth}
      \centering
      \includegraphics[width=1\textwidth]{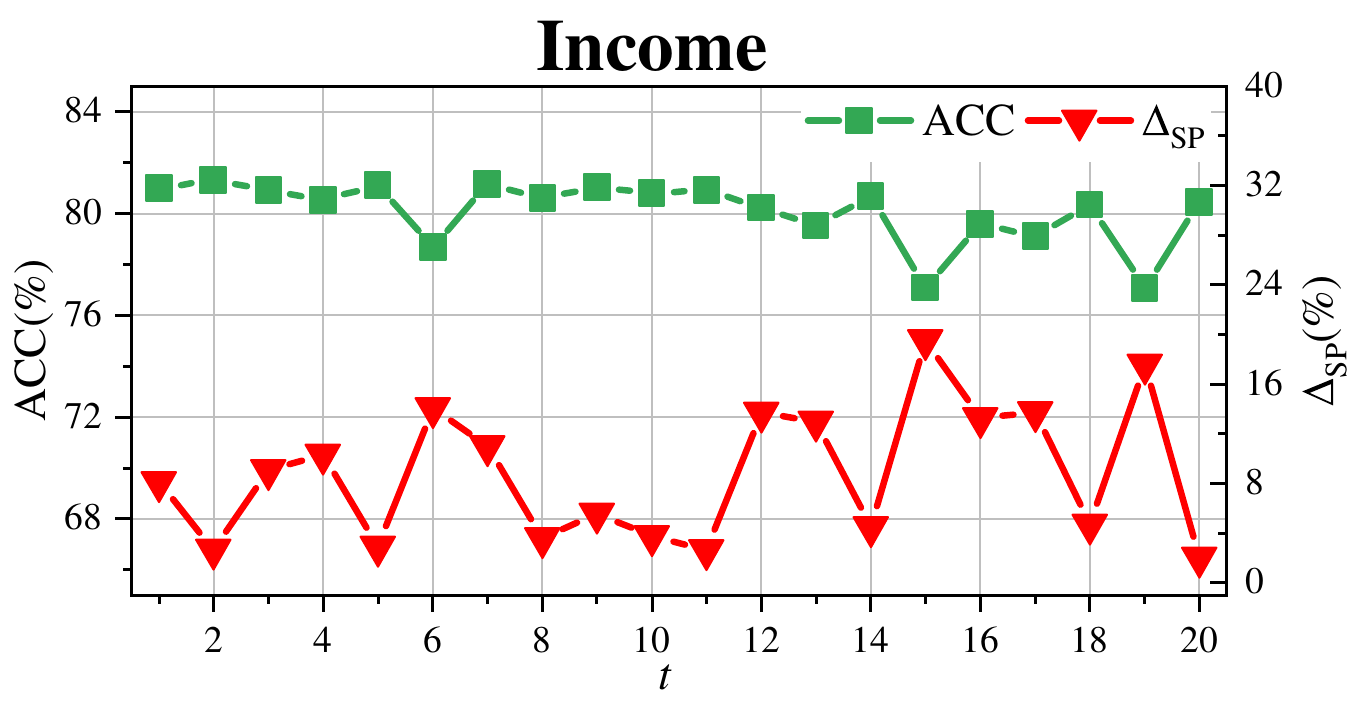}
    \end{minipage}
  }
  \vspace{-1em}
  \caption{The accuracy and $\Delta_{\text{SP}}$ of FairGT w.r.t. different parameter $t$ values.}
  \label{fig:parameter l}
  \vspace{-1em}
\end{figure*}

\subsection{Ablation Study}
FairGT as an fairness-aware GT achieves its objectives through two key components: eigenvector selection and $k$-hop sensitive information.
To comprehensively assess the contributions of these two elements, we conducted an ablation study.
Specifically, we aim to examine the extent to which eigenvector selection or the incorporation of $k$-hop sensitive information contributes to improving prediction fairness and accuracy.
In this analysis, we systematically remove each component independently to assess their individual impact.

We conduct a specific ablation analysis focused on the contribution of eigenvalue selection.
Because some GTs use Laplacian matrix to encode structural topology, we also consider eigenvalue selection of Laplacian matrix, represented as \textbf{Lap ST}.
The FairGT without eigenvalue selection is denoted as \textbf{w/o ST}.
Notably, when comparing \textbf{Lap ST} and \textbf{w/o ST}, FairGT consistently outperforms them in terms of accuracy and statistical parity.
The results shown in Table \ref{tab:ablation} reinforce the necessity of eigenvector selection within FairGT.
We denote the configuration as \textbf{w/o NF} (FairGT without node feature encoding).
We also consider $k$-hop neighborhood information based on adjacency matrix, represented as \textbf{Adj NF}.
The results shown in Table \ref{tab:ablation} provide compelling evidence of the contributions of the component.
When compared to the ablated versions, FairGT is consistently higher in terms of accuracy and lower in terms of $\Delta_{\text{SP}}$.
This observation underscores the pivotal role that $k$-hop sensitive information plays in concurrently improving the fairness of GTs' predictions and accuracy.

\begin{table}[H]
  \centering
  \footnotesize
  \tabcolsep=0.1cm
  \renewcommand{\arraystretch}{1.5}
  \begin{tabular}{lcccccc}
    \toprule
    \textbf{Dataset} & \textbf{Metric} & \textbf{FairGT} & \textbf{w/o ST} & \textbf{Lap ST} & \textbf{w/o NF} & \textbf{Adj NF} \\
    \midrule
    \textbf{NBA} &  ACC $\uparrow$        & \textbf{74.68}  & $58.23$ & $64.55$ & $63.29$ & $69.62$ \\
     & $\Delta_{\text{SP}}$ $\downarrow$  & \textbf{0.38}   & $6.21$  & $1.36$  & $2.42$  & $10.76$ \\
     \textbf{Bail} & ACC $\uparrow$       & \textbf{95.68}  & $93.96$ & $93.81$ & $90.76$ & $92.73$ \\
     & $\Delta_{\text{SP}}$ $\downarrow$  & \textbf{0.58}   & $6.42$  & $6.46$  & $5.87$  & $5.95$  \\
     \textbf{German} & ACC $\uparrow$     & \textbf{76.00}  & $75.19$ & $75.20$ & $70.80$ & $74.40$ \\
     & $\Delta_{\text{SP}}$ $\downarrow$  & \textbf{0.26}   & $2.29$  & $6.12$  & $1.30$  & $2.29$  \\
     \textbf{Credit} & ACC $\uparrow$     & \textbf{77.85}  & $75.20$ & $75.20$ & $75.20$ & $76.80$ \\
     & $\Delta_{\text{SP}}$ $\downarrow$  & \textbf{1.89}   & $2.29$  & $6.12$  & $7.40$  & $6.11$  \\
     \textbf{Income} & ACC $\uparrow$     & \textbf{81.30}  & $78.90$ & $80.65$ & $80.00$ & $81.00$ \\
     & $\Delta_{\text{SP}}$ $\downarrow$  & \textbf{2.66}   & $5.52$  & $4.28$  & $4.37$  & $5.42$  \\
    \bottomrule
  \end{tabular}
  \caption{Ablation study on different components of FairGT.}
  \label{tab:ablation}
  \vspace{-1em}
\end{table}

In summary, the ablation study results collectively emphasize the integral role played by eigenvector selection and $k$-hop sensitive information for improving fairness and performance of GTs.

\subsection{Parameter Analysis}
To comprehensively evaluate FairGT's performance, we investigate the impact of two crucial parameters: the number of eigenvectors $t$ and the number of Transformer layers $l$, conducting experiments across \textbf{NBA}, \textbf{Bail}, \textbf{German}, \textbf{Credit}, and \textbf{Income}.

With $l=2$, we select the numbers of feature vector in $\{1,2,3,4,5,...,20\}$, respectively.
Remarkably, $t$ differs across datasets to achieve peak performance due to the varied structural topologies of different networks.
The results are shown in Figure \ref{fig:parameter t}.
This performance fluctuation with $t$ increment suggests diverse effects of structural topology across different network types, influencing model performance diversely.
After comparing both fairness and performance metrics, we choose the best result as our final selection for the parameter $t$.
We select $t=5$ for \textbf{NBA}, $t=5$ for \textbf{Bail}, $t=11$ \textbf{German}, $t=2$ for \textbf{Credit}, and $t=2$ for \textbf{Income}.

Besides, we fix the best value of $t$ and change $l$ from $1$ to $5$ on each dataset.
The results are shown in Figure \ref{fig:parameter l}.
We select the best result as our final selection for the parameter $l$, after evaluating both fairness and performance metrics.
We set $l=2$ for \textbf{NBA}, $l=3$ for \textbf{Bail}, $l=3$ for \textbf{German}, $l=2$ for \textbf{Credit}, and $l=1$ for \textbf{Income}.

The results clearly indicate pronounced fluctuations in the outcomes for both parameters. The selection of values for $l$ and $t$ is crucial for FairGT.
Prudent choices of $l$ and $t$ are essential to obtain improved results in the experiment.
It is noteworthy that our choice might represent only a locally optimal solution, and hence conducting a sufficient number of experiments would be necessary to explore the entire parameter space. This aspect also suggests a potential avenue for refining the parameters of FairGT.

\section{Conclusion}
In this work, we have established a vital connection between graph information encoding and sensitive features to address the fairness issues in GTs.
We have proposed FairGT, an innovative approach dedicated to enhance the fairness of GTs.
FairGT is built on two fairness-aware graph information encoding: structural topology encoding and node feature encoding.
The designed components in FairGT collaborate to improve the fairness and performance of GTs. We also present theoretical analysis to prove the efficacy of the proposed approach.
In diverse real-world datasets, FairGT outperforms state-of-the-art baselines.
Despite achieving noteworthy results, our approach has limitations, as evidenced by pronounced fluctuations in outcomes when altering both parameters. This underscores the critical importance of selecting optimal values for $l$ and $t$.
In future work, we will further refine the efficiency of the FairGT approach and expand its applicability to situations with limited sensitive features.
We aspire to contribute to the ongoing progression of fairness-aware GTs and foster their broader adoption in real-world applications.

\section*{Appendix A}

\textbf{Lemma 1}
\textit{The dissimilarity between the distribution of original sensitive features and the distribution of $k$-hop neighbor sensitive features exhibits a pronounced correlation with the eigenvector corresponding to the largest magnitude eigenvalue of the adjacency matrix. Concurrently, the correlation with other eigenvectors diminishes exponentially.}

\proof
Assume $\mathbf{A} \in \mathbb{R}^{n \times n}$ is a adjacency matrix with real-valued entries.
The $k$-hop neighbor feature matrix is $\mathbf{A}^k \mathbf{H}$, and the $k$-hop neighbor sensitive features are $\mathbf{A}^k \mathbf{H}[:,s]$.
$|\lambda_1| > |\lambda_2| \geq ......  \geq |\lambda_n|$ are $n$ real eigenvalues, and $\mathbf{p}_i$ ($i \in \{1, 2, ......, n\}$) are corresponding eigenvectors.
The eigendecomposition of $\mathbf{A}$ can be written as $\mathbf{A} = \mathbf{P} \Lambda \mathbf{P}^\top$ with $\mathbf{P} = (\mathbf{p}_1, \mathbf{p}_2,......,\mathbf{p}_n)$, $\Vert \mathbf{p}_i \Vert = 1$ ($i \in \{1, 2, ......, n\}$) and $\Lambda = diag(\lambda_1, \lambda_2, ......, \lambda_n)$:
We assume $\alpha_i = \mathbf{H}[:,s]^ \top \mathbf{p}_i$.

We use cosine similarity to measure the dissimilarity. 

\begin{equation}
  \nonumber
  \begin{aligned}
  &cos(<\mathbf{p}_i, \mathbf{H}[:,s]>) \\
  =&\frac{\mathbf{H}[:,s]^\top \mathbf{p}_i}{\Vert \mathbf{H}[:,s] \Vert \Vert \mathbf{p}_i \Vert} \\
  =&\frac{\mathbf{H}[:,s]^\top \mathbf{p}_i}{\Vert \mathbf{H}[:,s] \Vert} \\
  =&\frac{\mathbf{H}[:,s]^\top \mathbf{p}_i}{\sqrt{\mathbf{H}[:,s]^\top \mathbf{H}[:,s]}} \\
  =&\frac{\mathbf{H}[:,s]^\top \mathbf{p}_i}{\sqrt{(\mathbf{P}^\top \mathbf{H}[:,s])^\top \mathbf{P}^\top \mathbf{H}[:,s]}} \\
  =&\frac{\mathbf{H}[:,s]^\top \mathbf{p}_i}{\sqrt{\sum_{j=1}^n(\mathbf{H}[:,s]^\top \mathbf{p}_j)^2}} \\
  =&\frac{\alpha_i}{\sqrt{\sum^n_{j=1}\alpha_j^2}}.
  \end{aligned}
\end{equation}

Then,
\begin{equation}
  \nonumber
  \begin{aligned}  
    &cos(<\mathbf{A}^k \mathbf{H}[:,s], \mathbf{H}[:,s]>) \\
  = &\frac{(\mathbf{A}^k \mathbf{H}[:,s])^\top \mathbf{H}[:,s]}{\Vert \mathbf{A}^k \mathbf{H}[:,s] \Vert \Vert \mathbf{H}[:,s] \Vert}\\
	= &\frac{(\mathbf{A}^k \mathbf{H}[:,s])^\top \mathbf{H}[:,s]}{\sqrt{(\mathbf{A}^k \mathbf{H}[:,s])^\top \mathbf{A}^k \mathbf{H}[:,s]} \sqrt{\mathbf{H}[:,s]^\top \mathbf{H}[:,s]}}\\
  = &\frac{(\mathbf{P} \Lambda^k \mathbf{P}^\top \mathbf{H}[:,s])^\top \mathbf{H}[:,s]}{\sqrt{(\mathbf{P} \Lambda^k \mathbf{P}^\top \mathbf{H}[:,s])^\top (\mathbf{P} \Lambda^k \mathbf{P}^\top \mathbf{H}[:,s])} \sqrt{\mathbf{H}[:,s]^\top \mathbf{H}[:,s]}}\\
  = &\frac{(\mathbf{P}^\top \mathbf{H}[:,s])^\top \Lambda^k (\mathbf{P}^\top \mathbf{H}[:,s])}{\sqrt{(\mathbf{P}^\top \mathbf{H}[:,s])^\top \Lambda^{2k} (\mathbf{P}^\top \mathbf{H}[:,s])} \sqrt{\mathbf{H}[:,s]^\top \mathbf{H}[:,s]}}\\
  = &\frac{\sum^n_{i=1}\alpha_i^2 \lambda_i^k}{\sqrt{\sum^n_{i=1}\alpha_i^2 \lambda_i^{2k}}\sqrt{\sum^n_{i=1}\alpha_i^2}}\\
	= &\frac{\alpha_1^2 + \sum^n_{i=2}\alpha_i^2 (\frac{\lambda_i}{\lambda_1})^k}{\sqrt{\alpha_1^2 + \sum^n_{i=2}\alpha_i^2 (\frac{\lambda_i}{\lambda_1})^{2k}}\sqrt{\sum^n_{i=1}\alpha_i^2}}.
  \end{aligned}
  \label{equ:eigenvector selection}
\end{equation}

Thus, the correlation between $\mathbf{p}_i$ and $cos(<\mathbf{A}^k \mathbf{H}[:,s], \mathbf{H}[:,s]>)$ is proportional to $(\frac{\lambda_i}{\lambda_1})^k$.
Because of $|\lambda_1| > |\lambda_i|$, the correlation decays exponentially.  

Specifically, when $k \rightarrow \infty$, we have:
\begin{equation}
  \nonumber
  \begin{aligned}
  &\lim_{k \to \infty} cos(<\mathbf{A}^k \mathbf{H}[:,s], \mathbf{H}[:,s]>) \\
  =&\lim_{k\rightarrow\infty}\frac{\alpha_1^2 + \sum^n_{i=2}\alpha_i^2 (\frac{\lambda_i}{\lambda_1})^k}{\sqrt{\alpha_1^2 + \sum^n_{i=2}\alpha_i^2 (\frac{\lambda_i}{\lambda_1})^{2k}}\sqrt{\sum^n_{i=1}\alpha_i^2}} \\
  =&\frac{\alpha_1}{\sqrt{\sum^n_{i=1}\alpha_i^2}} \\
  =&cos(<\mathbf{p}_1, \mathbf{H}[:,s]>).
  \end{aligned}
\end{equation}
\rightline{$\square$}

\section*{Appendix B}

\textbf{Lemma 2}
\textit{The sensitive feature distribution of $\mathbf{H}^{(k)}$ is the same as $\mathbf{H}$:}
\begin{equation}
  \mathbf{H}^{(k)}[i, s] = q^k\mathbf{H}[i, s],  
\end{equation}
\textit{where $i \in \{1,2,..., n\}$}, \textit{q denotes the number of nodes whose sensitive feature is 1.}
\proof The based case ($k=1$): 
$\mathbf{H}^{(1)} = \mathbf{A}_s\mathbf{H}$ and $\mathbf{H}[i, s] \in \{0, 1\}$.

Because 
\begin{equation}
  \nonumber
  \mathbf{H}^{(1)}[i,s] = \mathbf{A}_s[i,:]\mathbf{H}[:,s].
\end{equation}
\begin{equation}
  \nonumber
  \mathbf{A}_s[i,j]= \left\{ 
    \begin{aligned}
      1, \mathbf{H}[i,s] = \mathbf{H}[j, s],\\
      0, \mathbf{H}[i,s] \neq \mathbf{H}[j, s].
    \end{aligned}
  \right.
\end{equation}
\begin{equation}
  \nonumber
  \mathbf{A}_s[i,j]\mathbf{H}[j,s]= \left\{ 
    \begin{aligned}
      &0, \mathbf{H}[i,s]=0,\\
      &0, \mathbf{H}[i,s]=1,\mathbf{H}[j,s]=0,\\
      &1, \mathbf{H}[i,s]=\mathbf{H}[j,s]=1.
    \end{aligned}
  \right.
\end{equation}

Thus,
\begin{equation}
  \nonumber
  \mathbf{H}^{(1)}[i,s]= \left\{ 
    \begin{aligned}
      &0, \mathbf{H}[i,s]=0,\\
      &p, \mathbf{H}[i,s]=1.
    \end{aligned}
  \right.
\end{equation}

The conclusion is valid.

Inductive hypothesis ($k=r$): 
Assume that the conclusion holds when $k=r$, thus $\mathbf{H}^{(r)}[i,s] = q^r\mathbf{H}[i, s]$.
Thus, 
\begin{equation}
  \nonumber
  \mathbf{H}^{(1)}[i,s]= \left\{ 
    \begin{aligned}
      &0, \mathbf{H}^{(r-1)}[i,s]=0,\\
      &q^r, \mathbf{H}^{(r-1)}[i,s]=q^{r-1}.
    \end{aligned}
  \right.
\end{equation}

Inductive Step ($k=r+1$):

Because $\mathbf{H}^{(r+1)}[i,s] = \mathbf{A}_s[i,:]\mathbf{H}^{(r)}[:,s]$.
\begin{equation}
  \nonumber
  \mathbf{A}_s[i,j]\mathbf{H}^{(r+1)}[j,s]= \left\{ 
    \begin{aligned}
      &0,   &&\mathbf{H}^{(q)}[i,s]=0,\\
      &0,   &&\mathbf{H}^{(q)}[i,s]=q^r,\\
      &     &&\mathbf{H}^{(q)}[j,s]=0,\\
      &q^r, &&\mathbf{H}^{(q)}[i,s]=q^r,\\
      &     &&\mathbf{H}^{(q)}[j,s]=q^r.
    \end{aligned}
  \right.
\end{equation}

Thus,
\begin{equation}
  \nonumber
  \mathbf{H}^{(r+1)}[i,s]= \left\{ 
    \begin{aligned}
      &0, \mathbf{H}^{(r)}[i,s]=0,\\
      &q^{(r+1)}, \mathbf{H}^{(r)}[i,s]=q^r.
    \end{aligned}
  \right.
\end{equation}

The conclusion is confirmed in the inductive step.

Combining the foundational step, the inductive hypothesis, and the inductive step, we show that this mathematical statement holds for all positive integer $k$. 
\rightline{$\square$}

\section*{Appendix C}

More details of experimental results.
\begin{table}[H]
  \centering
  \scriptsize
  \tabcolsep=0.05cm
  \begin{tabular}{lcccccccccc}
    \toprule
    \multirow{2}[2]{*}{\textbf{Methods}} & \multicolumn{2}{c}{\textbf{NBA}} & \multicolumn{2}{c}{\textbf{Bail}} & \multicolumn{2}{c}{\textbf{German}} & \multicolumn{2}{c}{\textbf{Credit}} & \multicolumn{2}{c}{\textbf{Income}} \\
    \cmidrule{2-11}
    & F1 $\uparrow$ & AUC $\uparrow$ & F1 $\uparrow$ & AUC $\uparrow$ & F1 $\uparrow$ & AUC $\uparrow$ & F1 $\uparrow$ & AUC $\uparrow$ & F1 $\uparrow$ & AUC $\uparrow$\\
    \midrule
    \textbf{GCN}        & $74.42$ & $71.10$ & $79.15$ & $87.43$ & $81.61$ & $62.81$ & $78.83$ & $69.45$ & $51.02$ & \textbf{77.33}\\
    \textbf{GCNII}      & $76.09$ & $77.63$ & $89.43$ & $94.13$ & $82.97$ & $53.16$ & $84.46$ & $61.22$ & $41.71$ & $67.94$\\
    \textbf{GAT}        & $75.00$ & $73.97$ & $90.65$ & $96.37$ & $82.74$ & $57.55$ & $76.40$ & $64.86$ & $35.77$ & $61.21$ \\
    \midrule
    \textbf{FairGNN}    & $69.48$ & $77.84$ & $77.50$ & $87.36$ & $82.01$ & $67.35$ & $77.79$ & \textbf{69.73} & $50.93$ & $77.24$\\
    \textbf{NIFTY}      & $71.91$ & $68.85$ & $76.40$ & $78.20$ & $82.75$ & $68.78$ & $84.08$ & $68.30$ & $41.27$ & $70.21$\\
    \textbf{BIND}       & $67.41$ & $62.82$ & $85.62$ & $93.93$ & $81.93$ & $71.88$ & $78.04$ & $69.54$ & $50.55$ & $76.89$\\
    \midrule
    \textbf{GTrans} & $76.92$ & $74.81$ & $91.56$ & $97.32$ & $84.24$ & $75.54$ & $80.75$ & $68.13$ & $38.90$ & $63.60$\\
    \textbf{SAN}        & $73.42$ & $75.51$ & $93.07$ & $97.84$ & $83.37$ & $76.79$ & $80.46$ & $63.30$ & $32.09$ & $63.29$\\
    \textbf{NAG} & $76.09$ & $78.91$ & $90.67$ & $97.46$ & $84.42$ & $71.44$ & $86.91$ & $63.30$ & $40.20$ & $63.60$\\
    \midrule
    \textbf{FairGT}     & \textbf{78.16} & \textbf{81.22} & \textbf{94.46} & \textbf{99.07} & \textbf{84.54} & \textbf{78.32} & \textbf{87.19} & $69.38$ & \textbf{51.46} & $76.01$\\
    \bottomrule
  \end{tabular}
  \caption{Comparison of performance (F1 and AUC) in percentage (\%). $\uparrow$ denotes the larger, the better. The best results are bold-faced.}
  \vspace{-1em}
  \label{tab:result}
\end{table}

\bibliographystyle{named}
\bibliography{FairGT}

\end{document}